\pdfoutput=1

\documentclass[11pt]{article}

\usepackage[final]{acl}

\usepackage{times}
\usepackage{latexsym}

\usepackage[T1]{fontenc}

\usepackage[utf8]{inputenc}

\usepackage{microtype}

\usepackage{inconsolata}

\usepackage{graphicx}

\usepackage{times}
\usepackage{latexsym}
\usepackage{caption}
\usepackage{subcaption}
\usepackage{amsmath}
\usepackage{amssymb}
\usepackage{amsfonts}
\usepackage{subfloat}
\usepackage{xcolor}
\usepackage{siunitx}

\usepackage[T1]{fontenc}

\usepackage[utf8]{inputenc}

\usepackage{microtype}

\usepackage{inconsolata}

\usepackage{graphicx}
\usepackage{todonotes}

\usepackage{microtype}
\usepackage{graphicx}
\usepackage{tabularx}
\usepackage{xspace}
\usepackage{multirow}
\usepackage{enumitem}
\usepackage{booktabs}
\usepackage{adjustbox}

\usepackage{tikz}
\usetikzlibrary{bayesnet}

\usepackage{algpseudocode}

\newcommand{\camerareadytext}[1]{\xspace}

\newcommand{\sref}[1]{\S\ref{#1}}
\newcommand{\fref}[1]{Figure~\ref{#1}}

\newcommand{\nutmeg}{\textsc{Nutmeg}\xspace}
\newcommand{\mace}{\textsc{Mace}\xspace}

\newcommand{\myparagraph}[1]{\noindent \textbf{#1}}

\title{NUTMEG: Separating Signal From Noise in Annotator Disagreement}

\author{Jonathan Ivey \\
  University of Arkansas \\
  \texttt{jwi001@uark.edu} \\\And
  Susan Gauch \\
  University of Arkansas \\
  \texttt{sgauch@uark.edu} \\\And
  David Jurgens \\
  University of Michigan \\
  \texttt{jurgens@umich.edu} \\}

\begin{document}
\maketitle
\begin{abstract}

NLP models often rely on human-labeled data for training and evaluation. Many approaches crowdsource this data from a large number of annotators with varying skills, backgrounds, and motivations, resulting in conflicting annotations. These conflicts have traditionally been resolved by aggregation methods that assume disagreements are errors. Recent work has argued that for many tasks annotators may have genuine disagreements and that variation should be treated as signal rather than noise. However, few models separate signal and noise in annotator disagreement. In this work, we introduce \nutmeg, a new Bayesian model that incorporates information about annotator backgrounds to remove noisy annotations from human-labeled training data while preserving systematic disagreements. Using synthetic data, we show that \nutmeg is more effective at recovering ground-truth from annotations with systematic disagreement than traditional aggregation methods. We provide further analysis characterizing how differences in subpopulation sizes, rates of disagreement, and rates of spam affect the performance of our model. Finally, we demonstrate that downstream models trained on \nutmeg-aggregated data  significantly outperform models trained on data from traditionally aggregation methods. Our results highlight the importance of accounting for both annotator competence and systematic disagreements when training on human-labeled data.

\end{abstract}

\section{Introduction}

\begin{figure}[t]
    \centering
    \includegraphics[width=\linewidth]{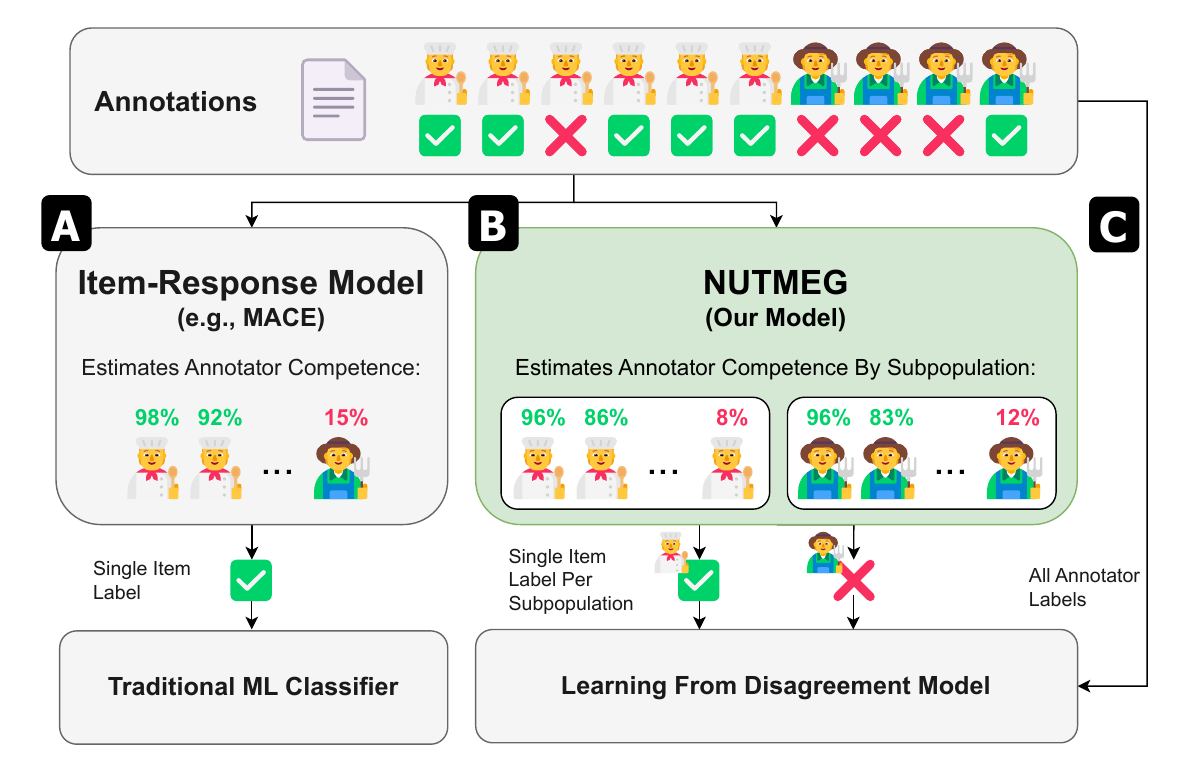}
    \caption{When aggregating annotators' labels, (A) traditional item response models like \mace \citep{hovy-etal-2013-learning} ignore meaningful label variation in subpopulations to produce a single label, while (C) learning from disagreement models take all annotations as input, ignoring potential spam labels. Our approach (B), \nutmeg, extends the item response paradigm to infer labels per subpopulation, which can provide  more accurate inputs when learning from disagreement. }
    \label{fig:intro-fig}
\end{figure}

NLP is largely dependent on labeled data to train and evaluate models. Typically, labels are generated by humans through an annotation process that involves aggregating the judgments of multiple individuals \citep{snow-etal-2008-cheap, nowak_crowdsourcing_2010,zheng2017truth}. Given the cost of experts, many approaches opt for crowdsourcing labels from a large number of annotators who have varying levels of training and expertise in the task. In this setting, some labels are assumed to be errors due to mistakes, ambiguity in the task, or even adversarial behavior by annotators \citep{hsueh-etal-2009-data, aroyo_welty_2014, jagabathula_identifying_2017}. As a result, there are many approaches for estimating an item's true label the from potentially-noisy collective annotations \citep{whitehill_whose_2009, liu_variational_2012,zheng2017truth,paun-etal-2018-comparing, goh_crowdlab_2023, bernhardt_active_2022}. While these models are effective, they assume that any deviation from the consensus label is a mistake. However, recent work has established that annotators from specific backgrounds may systematically differ in their judgments on an item, particularly for subjective tasks  \citep{larimore-etal-2021-reconsidering, sap-etal-2022-annotators, pei-jurgens-2023-annotator, wan_everyones_2023, mostafazadeh-davani-etal-2024-d3code}. As a result, annotator groups who systematically disagree with the majority label are likely lost in aggregation. Here, we introduce a new Bayesian model for inferring ground truth labels that incorporates annotator backgrounds and allows for identifying systematic disagreement in labels (Figure \ref{fig:intro-fig}).

Annotators disagree and two strands of research have proposed approaches to resolve these disagreements. One strand has framed the disagreement resolution as an unsupervised learning problem where a model simultaneously learns the probable ground truth label while also learning which labelers are more likely to give accurate answers \citep{zheng2017truth}. Approaches such as \mace \citep{hovy-etal-2013-learning} use Bayesian models to infer a single ground truth label per item, which is suitable for training most machine learning models. However, there are many subjective tasks, such as detecting hate speech, where the assumption of a single label can cause these models to ignore valid disagreements.

In contrast, a more recent strand has noted that some disagreements are meaningful and proposed new machine learning methods for \textit{Learning from Disagreement} \citep{uma_learning_2021}; such methods learn to predict from the original disaggregated data. While this latter branch is effective at incorporating diverse views---e.g., how different groups might view the same item---such models likely overweigh non-systematic disagreement, such as those due to mistakes or adversarial behavior. 

Here, we introduce a Bayesian model for learning ground truth labels that is able to model systematic variation between subpopulations within the annotators. Our approach, \nutmeg (\textbf{N}uanced \textbf{U}nderstanding of anno\textbf{T}ation by \textbf{M}ultipl\textbf{E} \textbf{G}roups) estimates annotator competence and infers per-subpopulation labels for each item. In experiments on synthetic data, we demonstrate that (i) \nutmeg can accurately recover distinct labels for each subpopulation when they differ, while still recognizing when annotators are spamming and (ii) \nutmeg is effective even for small numbers of annotations per subpopulation, making it readily amenable to use with crowdsourcing. Finally, in experiments with real data labeled with demographics, we show that by first reducing noise with \nutmeg, we can use subpopulation labels with learning from disagreement models to make more accurate predictions. We release \nutmeg and the synthetic data generation framework at {\small{https://github.com/jonathanivey/NUTMEG}}.

\section{Modeling Annotator Disagreements}

Annotators may disagree with each other for a variety of reasons---valid or not---and multiple branches of research have focused on understanding or resolving these disagreements to improve machine learning performance.

\myparagraph{Modeling Annotator Backgrounds}
An individual's background (e.g., demographics, occupation) is known to systematically influence their annotation behavior, leading to disagreements in labeling \citep{Lerner2024WhosePD,pei-jurgens-2023-annotator}. 
While not focused on resolving these disagreements, recent work in NLP has focused on understanding how much of the disagreement can be attributed to an annotator's background; for example, showing that conservative annotators are less likely to rate anti-Black language as toxic \citep{sap-etal-2022-annotators}.
While earlier annotated data rarely included information about the annotators, more recent work has called for a responsible collection of this data \citep{Santy2023NLPositionalityCD,davani2022dealing}, particularly for improving models by including diverse viewpoints \citep[e.g.,][]{fleisig-etal-2023-majority,Orlikowski2023TheEF} and identifying biases in LLM behaviors \citep[e.g.,][]{Santy2023NLPositionalityCD,Deng2023YouAW}. Our work fills a key gap \camerareadytext{in this line of research} by showing how to incorporate systematic diversity in groups' ratings in modeling while still accounting for noise and mistakes during the annotation process.

\myparagraph{Inferring Annotator Competence}
Prior work has identified differences in annotator labels as label-noise and attempted to reduce it prior to model training \citep{dawid_maximum_1979}. Many of these methods use unsupervised probabilistic models of annotator behavior to identify incorrect labels, also known as spam, and estimate annotator competence \citep{whitehill_whose_2009, liu_variational_2012, hovy-etal-2013-learning, paun-etal-2018-comparing}. More recent models have used information from classifiers for the same goal of estimating annotator competence or reducing \camerareadytext{label} noise \citep{goh_crowdlab_2023, bernhardt_active_2022}.

These methods rely on the assumption that disagreements between annotators indicate errors or a lack of competence; however, for many tasks there can be genuine disagreements between annotators \citep{plank-etal-2014-linguistically}. In this work, we create a new model of annotator competence that retains genuine disagreements between annotators while also reducing label-noise.

\myparagraph{Task Subjectivity}
There are many causes for annotator disagreements, also known as human label variation, including expertise, item-difficulty, and motivation \citep{plank-2022-problem, fleisig-etal-2024-perspectivist}, but a particularly important cause is how an annotator's background interacts with the task objectives. Previous work has shown that in many subjective NLP tasks, like detecting offensiveness, politeness, and toxicity, annotator disagreements are correlated with backgrounds and social variables such as gender, race, age, and region \citep{larimore-etal-2021-reconsidering, sap-etal-2022-annotators, pei-jurgens-2023-annotator, wan_everyones_2023, mostafazadeh-davani-etal-2024-d3code}. This relationship is particularly important because current models of annotator competence treat consensus as correct---and disagreement as error---and as a result, valid label disagreements by minority subpopulations risk being omitted. %

In this work, we introduce a new model to estimate different truth values for each relevant subpopulation for each item in a dataset. This multiple-truth modeling approach allows us to pass disagreement---in connection with its systemic causes---to downstream modeling applications and reduce the risk of omitting meaningful variation in labeling decisions.

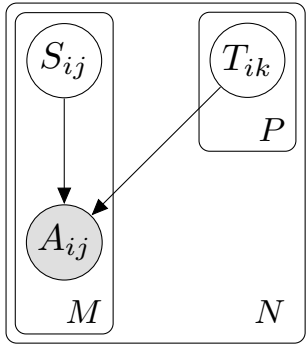
\begin{figure}[t]
    \centering
    \begin{tikzpicture}[scale=1.4, transform shape]
        \node[latent] (T) {$T_{ik}$};
        \node[obs, left=of T, yshift=-1.725cm] (A) {$A_{ij}$};
        \node[latent, above=of A] (S) {$S_{ij}$};
        
        \edge {S} {A};
        \edge {T} {A};
        
        \plate[inner sep=0.25cm] {item} {(S) (A) (T)} {$N$};
        \plate {annotator} {(A) (S)} {$M$};
        \plate {subpopulation} {(T)} {$P$};
    \end{tikzpicture}
    
    \caption{Plate diagram of \nutmeg. Annotator $j$ from subpopulation $k$ produces label $A_{ij}$ on instance $i$. The label choice depends on the instance's true label for subpopulation $k$, $T_{ik}$, and whether $j$ is spamming on $i$, modeled by binary variable $S_{ij}$. $N =|instances|$, $M = |annotators|$, and $P = |subpopulations|$.}
    \label{fig:plate_diagram}
\end{figure}

\myparagraph{Learning from Disagreement}
When given data with conflicting judgments, one line of research known as \textit{Learning from Disagreement} has proposed treating disagreement as signal rather than noise. The simplest approach to treating disagreement as a signal is to not aggregate annotations at all and instead use the full distribution of responses for each item as the desired output of a model \citep{uma_learning_2021}.

Other approaches choose to model every annotator response by training multi-task models \citep{davani-etal-2022-dealing, mokhberian-etal-2024-capturing} or training on every annotator-item pair \citep{gordon_jury_2022, fleisig-etal-2023-majority, weerasooriya-etal-2023-disagreement}. These methods are effective at incorporating annotator diversity, but they do not account for other causes of disagreement like mistakes and adversarial behavior that may reduce model performance.

Our work introduces a complementary method to enable noise reduction while retaining disagreement, which can be used in combination with these Learning from Disagreement approaches to reduce noise prior to training. %
Limited work has focused on both reducing noise and retaining disagreement in human labels. \citet{weber-genzel-etal-2024-varierr} designed models to separate annotation errors from valid disagreements; however, their work is designed for natural language inference and relies on a more-involved setting that collects and evaluates explanations by annotators to remove errors. Here, we use subpopulation information to identify systemic disagreement, and our method, which does not require the collection of annotator explanations, can be applied to existing datasets.

The most similarly themed work to ours are the CrowdTruth 2.0 metrics \citep{dumitrache_crowdtruth_2018}, which are designed to capture media unit quality, worker quality, and annotation quality. Though CrowdTruth 2.0 does model disagreement, it does not model the origins of disagreement or produce training labels for downstream tasks. In this work, we use relevant information about annotators, through subpopulation divisions, to model the causes of systematic disagreement, allowing downstream stakeholders to attribute disagreement and train models tailored to different use-cases.

\section{Methods}

We introduce \nutmeg, a Bayesian model that estimates annotator competence and predicts item labels while retaining subpopulation-level disagreement (\fref{fig:plate_diagram}). Our approach builds upon  Bayesian model designs for items' and annotators' variation \citep[e.g.,][]{paun-etal-2018-comparing}. \nutmeg most closely resembles \mace \citep{hovy-etal-2013-learning} which attempts to simultaneously learn the spamming rates of annotators and the item's likely label.

One key assumption of prior Bayesian models is that there exists a single correct label for each item agreed upon by annotators. In \nutmeg, we relax this assumption so that there exists a single correct label \textit{for each subpopulation} that is always given by an annotator in that subpopulation when they try to. While real data likely has additional within-group variation, this simplifying assumption allows our model to focus on capturing systematic variation that is most relevant to downstream applications.

To incorporate subpopulation identity, the generative step of our model works as follows (also shown in \ref{fig:generative_step}): First, for each instance $i$ and subpopulation $k$, we sample the true subpopulation label $T_{ik}$ from a uniform prior. Then, for each annotator $j$, we sample a binary variable $S_{ij}$ from a Bernoulli distribution with parameter $1 - \theta_j$. $S_{ij}$ represents whether annotator $j$ is spamming on instance $i$. If the annotator is not spamming,\footnote{Here, we follow common notation and refer to a deviation from the correct label as ``spam.'' However, this label category reflects any type of disagreement, adversarial or not.} then they use the \textit{true label of their subpopulation} to produce annotation $A_{ij}$. Otherwise, their annotation $A_{ij}$ is sampled from a multinomial with parameter $\xi_j$. 
The parameter $\theta_j$ represents the probability that annotator $j$ is not spamming on a given instance; this is a measure of their competence. The parameter $\xi_j$ represents annotator $j$'s individual behavior when they are spamming, which could produce the correct label, but only by chance.

To fit the model, we use Variational-Bayes training to maximize the probability of the observed data:
\begin{align*}
     P(A; \theta, \xi) &= 
     \sum_{T, S} \Bigg[
     \prod_{i=1}^{N} \prod_{k=1}^{P} P(T_{ik}) \cdot \\
     &\quad \prod_{j=1}^{M} P(S_{ij};\theta_j) \cdot 
     P(A_{ij} | S_{ij}, T_{ik}; \xi_j) \Bigg]
\end{align*}
We train with symmetric Beta priors with parameters of 0.5 on $\theta_j$ and symmetric Dirichlet priors on $\xi_j$. As identified in \mace, these priors model the extremes of behavior common in annotation (i.e., either an annotator often gives the correct label or they rarely give the correct label). However, \nutmeg also supports adjusting these priors should an end-user desire a more informed prior. 

Because not every subpopulation is guaranteed to label every item, \nutmeg must handle instances where there is an unobserved subpopulation $k_{x}$ for item $i_{x}$. In those cases, we calculate the estimated label for the observed subpopulations of $i_{x}$, then we identify the set of items that have the same estimated labels for those subpopulations and contain annotations from $k_{x}$. Finally, we take the average of the posterior probabilities for the items in this set to estimate the posterior of $T_{i_xk_x}$. For this process, we make the simplifying assumption that observed labels from different subpopulations are independent. Though this is often not the case, it allows us to use a larger number of items to estimate unobserved instances. Future work could explore more robust methods for estimating items' labels for unobserved subpopulations.
We note that \nutmeg does not require the use of these imputed subpopulation truths, and for Experiment 3 (\sref{sec:Exp4}), we choose not to estimate unobserved samples to reduce the risk of introducing additional label noise.

\nutmeg requires that each annotator be associated with a subpopulation label. Given the NLP communities recent recognition of the importance of collecting information about the annotators themselves \cite{Lerner2024WhosePD,Santy2023NLPositionalityCD,mihalcea2025ai} and the recent uptick in the creation of such datasets \citep[e.g.,][]{kumar2021designing,sap-etal-2022-annotators,pei-jurgens-2023-annotator}, we believe that data on annotators will be increasingly important and available. However, we note that \nutmeg does not require that these subpopulation labels be derived from questionnaires; alternative work in NLP has proposed grouping annotators based on behavior \cite[e.g.,][]{vitsakis-etal-2024-voices} and these inferred groups could serve as imputed subgroup labels for modeling their systematic behavior with \nutmeg.

\begin{figure}[t]
    \centering
    \begin{align*}
        & \text{for } i = 1 \dots N: \\
        & \hspace{0.5cm}\text{for } k = 1 \dots P: \\
        & \hspace{1cm} T_{ik} \sim \text{Uniform} \\
        & \hspace{0.5cm}\text{for } j = 1 \dots M: \\
        & \hspace{1cm} S_{ij} \sim \text{Bernoulli(} 1-\theta_j \text{)} \\
        & \hspace{1cm} \text{if } S_{ij} = 0: \\
        & \hspace{1.5cm} A_{ij} = T_{ik} \\
        & \hspace{1cm} \text{else}: \\
        & \hspace{1.5cm} A_{ij} \sim \text{Multinomial(} \xi_j \text{)}
    \end{align*}
    \caption{The generative process for \nutmeg. See the text for a full description of variables.}
    \label{fig:generative_step}
\end{figure}

\section{Experiment 1: Synthetic Data} 
\label{sec:Exp1}

\begin{figure*}[t]
    \centering
    \includegraphics[width=0.9\linewidth]{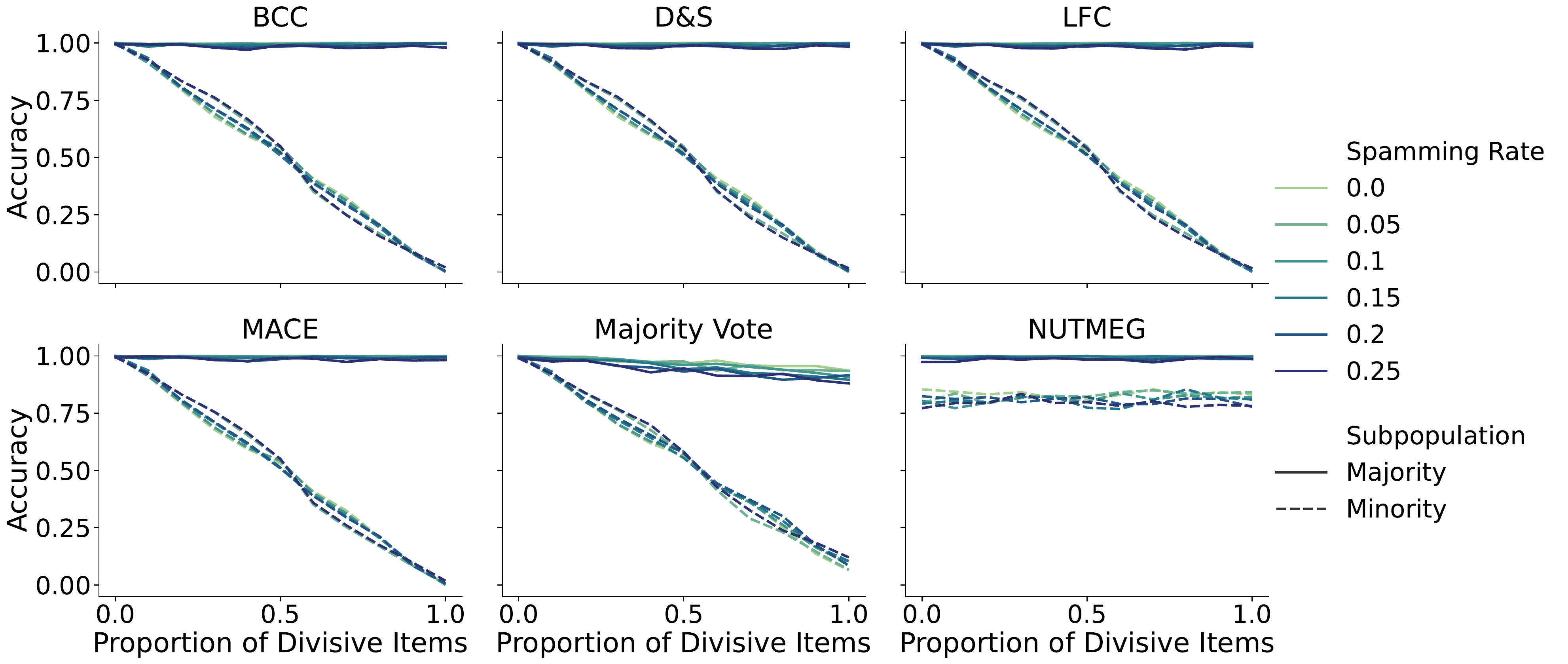}
    \caption{As the rate of disagreement increases between majority and minority populations for the subset of items, traditional models are increasingly less accurate at correctly inferring the subpopulation's true label (dashed line), while still being accurate for the majority subpopulation (solid line). By contrast, \nutmeg can effectively estimate the opinions of both the majority and minority subpopulations with varying rates of spam (shown with color).}
    \label{fig:subpop_disagreement}
\end{figure*}

To evaluate how effective \nutmeg is at reducing noise and recovering ground truth from multiple subpopulations, we first evaluate performance using synthetic data that precisely simulates annotator behavior in a setting with systematic disagreement. Using synthetic data allows us to compare \nutmeg's estimates to the true opinions of annotators that are not available in real annotated data. 

\subsection{Experimental Setup}

To generate the synthetic data, we first create a set of 150 annotators and assign them randomly to one of two subpopulations, a majority and a minority, using an 80\% and 20\% split for this experiment. We then assign each annotator a spamming rate in [0,1]; these rates represent the probability that an annotator ignores their subpopulation's true label when labeling an item. %
The mean spamming rate across individuals indicates the overall level of spam in the data. After generating our annotators, we create a set of 500 items with two possible labels. We designate a proportion of the items as \textit{divisive} according to a global divisiveness rate. For divisive items, the subpopulations will hold different true opinions, and for non-divisive items, they  will hold the same true opinion. The divisiveness rate indicates the level of systematic disagreement in the data.

To simulate the annotation process for an item, we first decide whether each annotator is spamming based on their competence score. Spammers assign labels to the item randomly, while non-spammers provide the true opinion of their subpopulation. Finally, to simulate the availability of crowdsourced annotations we randomly sample from this dataset so that each item has 5 annotations and each annotator labels more than 20 items (average of 16.67 items per annotator). We will release this synthetic data and evaluation framework for future research on modeling subpopulation variation in annotation.

To evaluate how effective \nutmeg is at recovering ground truth, we use the above procedure to generate multiple synthetic datasets with different divisiveness rates varying from 0 to 1 and global spamming rates varying from 0 to 0.25. 
We compare performance against five models for estimating ground truth to show the effect of systematic variation by subpopulation. We include a majority vote, the original \citet{dawid_maximum_1979} model (D\&S), and its extension \mace, which is the closest  comparison to our model. We also follow the model recommendations from the large survey by \citet{zheng2017truth} and include the Learning from Crowds \citep[LFC; ][]{raykar2010learning} and Bayesian Classifier Combination \citep[BCC;][]{kim2012bayesian}, which perform best for the type of nominal data used in our experiments. We fit all five models on each dataset and calculate an accuracy score by comparing a model's estimates for each subpopulation to its true label in the dataset.

\begin{figure}[t]
    \centering
    \includegraphics[width=\linewidth]{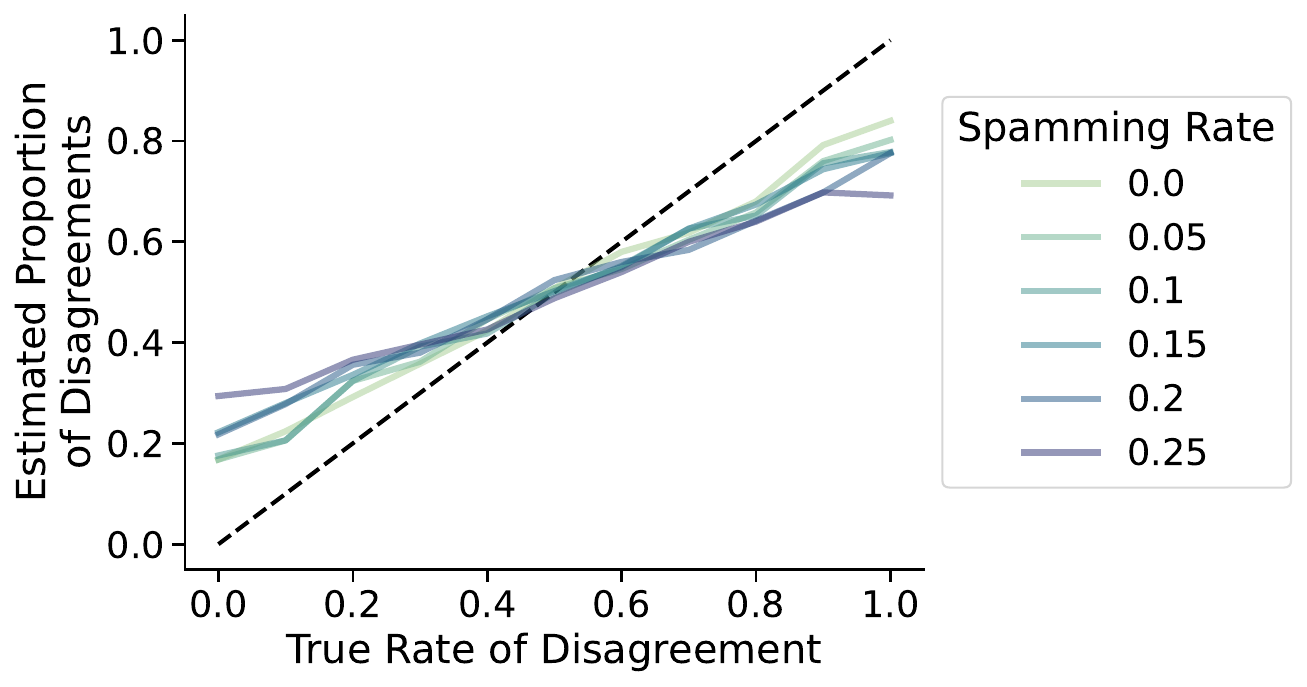}
    \caption{\nutmeg can effectively differentiate genuine disagreement from spam annotations. As the rate of spam increases, \nutmeg's estimate of the disagreement rate diverges from the true rate, especially at the extremes. The dotted line indicates perfect predictions.}
    \label{fig:disagreement_estimate}
\end{figure}

\subsection{Results}

We find that as the rate of systematic disagreement increases,  \nutmeg correctly identifies the true label for both the majority and minority subpopulations (\fref{fig:subpop_disagreement}), despite the minority having four times less available data. Importantly, \nutmeg makes these improvements without significantly reducing its ability to estimate the majority opinion. In contrast, while the other methods are accurate for recovering the majority's true label, they are increasingly inaccurate for the minority subpopulation's true label as the rate of divisiveness increases (as expected).

Note that while \nutmeg is able to recover most of the true labels, the accuracy for the minority subpopulation is still lower. This gap is primarily due to data sparsity. For some items, too few minority annotators may be assigned to accurately distinguish meaningful disagreement from spam. Further, as spam rates increase from 0 to 25\%, we see an average 4.22\% drop in minority accuracy. While a 25\% spam rate is likely on the high side, \nutmeg's overall accuracy is still high regardless of how often the subpopulation disagrees. As experimenters may not know how often a particular group might disagree, this performance trend suggests \nutmeg can be accurately deployed even when divisive items are relatively rare.

\myparagraph{How good is \nutmeg at distinguishing systematic disagreement from spam?}
In our simulations, all annotators are capable of spamming and thus, not all divergent labels by annotators in the minority subpopulation are meaningful.  To assess whether \nutmeg recognizes these labels as spam, we compare \nutmeg's estimated proportion of disagreements to the true global rate of divisiveness. This gives an indication of how accurately \nutmeg can differentiate between genuine systemic disagreement and spam. 

We find that \nutmeg can effectively differentiate genuine disagreement from spam annotations, as shown in \fref{fig:disagreement_estimate}. However, as the rate of disagreement approaches the high and low extremes, \nutmeg's estimate of the rate of disagreement diverges from the true rate. At especially high rates of disagreement, the model underestimates and at especially low rates of disagreement, the model overestimates. We believe that this divergence is caused by the increased effect that contrary evidence has at extreme rates of disagreement. Even a single annotation that diverges from expectations may bring the model's estimates away from extremes, and this effect is exacerbated by spam, which provides increased contrary evidence. 

We further find that \nutmeg is effective at assessing annotator competence. The average Pearson's correlation between \nutmeg's estimate of annotator competence $\theta_j$ and the annotator's true competence across all runs is $0.81$. By comparison, \mace's average correlation is $0.58$. These correlations do not significantly differ for members of the majority or minority subpopulations, which illustrates how using traditional item-response models can lead a practitioner to erroneously conclude that they have low-quality annotators when in fact they have high-quality annotators and systematic disagreement.

\section{Experiment 2: Subpopulation Size}
\label{sec:Exp2}

Practitioners labeling data often have limited control of how many annotators in specific groups are present in the annotator pool. Experiment 1 (\sref{sec:Exp1}) showed a decrease in accuracy for estimating the minority opinion as the rate of spam increases. %
This result leads to a natural question: how much data is enough to accurately represent a minority subpopulation given different rates of spam?

\subsection{Experimental Setup}

To test for the effect of minority subpopulation size during annotation, we repeat our synthetic data generation procedure from \sref{sec:Exp1}, but fix the global spam rate of 0.1 and a global divisiveness rate of 0.2. We then vary the size of the minority subpopulation from 10\% to 50\% and the total number of annotators sampled for each item from 3 to 15. Note that because annotators are randomly assigned to items, for settings with few annotations per item, some items may not receive any annotators by a person of the minority subpopulation, which mirrors real world settings. Finally, we run \nutmeg on the datasets and calculate an accuracy score to compare its estimates to the true opinions of the minority subpopulation. 

\subsection{Results}

\begin{figure}[t]
    \centering
    \includegraphics[width=\linewidth]{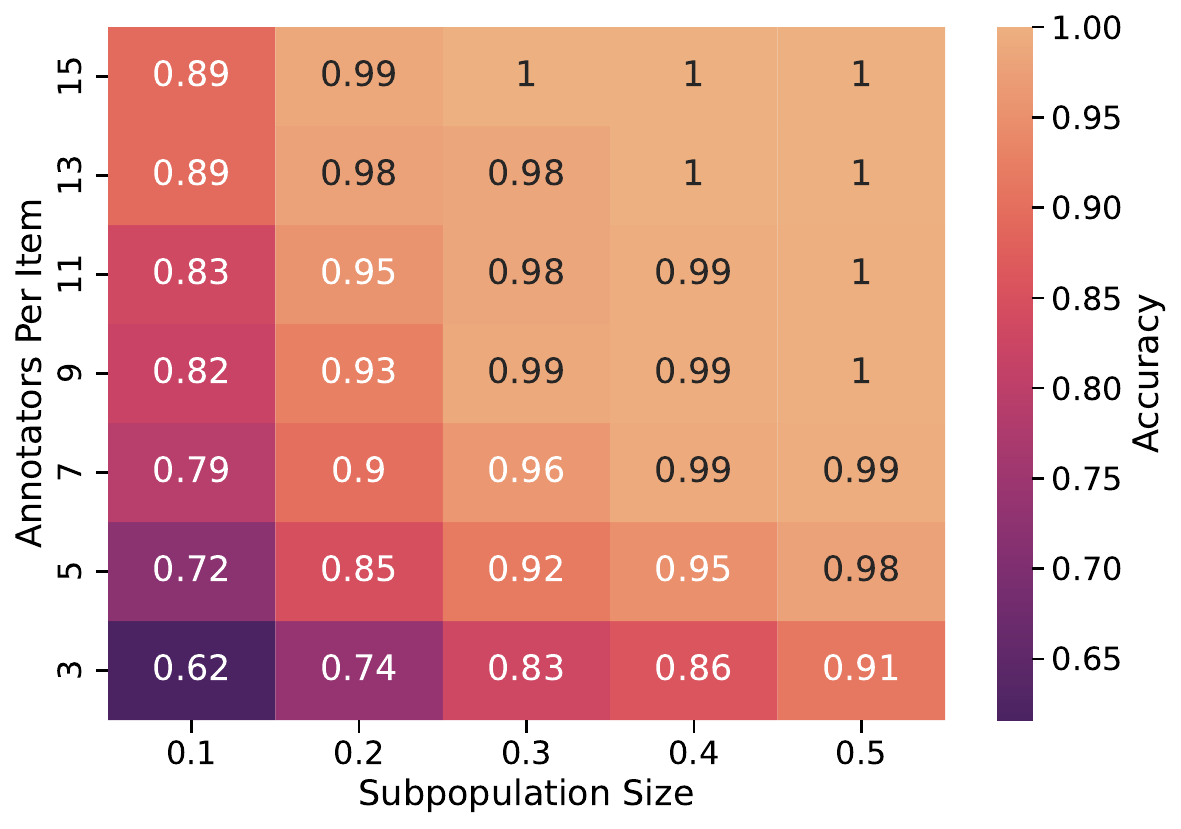}
    \caption{As the size of a subpopulation decreases, \nutmeg needs more annotations to ensure that the subpopulation is sufficiently represented and estimate its true opinions. This result highlights the importance of data collection strategies for representing minority subpopulations.}
    \label{fig:subpop_size}
\end{figure}

We find that as the size of a subpopulation in a dataset gets smaller, \nutmeg requires a much larger number of annotations per item to maintain the same level of accuracy (\fref{fig:subpop_size}). With a subpopulation proportion of 0.3 (equivalent to 45 annotators), \nutmeg only requires 5 annotations to achieve 92\% accuracy, but for a subpopulation proportion of 0.1 (equivalent to 15 annotators), \nutmeg would need more than 15 annotations per item to achieve the same performance. This result shows the importance of intentional data collection methods when representing the opinions of small subpopulations. If new datasets need an accurate estimate for multiple subpopulations, they should focus on having sufficient coverage for each subpopulation rather than simply increasing the total number of annotations collected. Alternatively, training sets can choose not to calculate estimates for sets items without sufficient labels from small subpopulations, which is the approach that we take for model training in \sref{sec:Exp4}.

\section{Experiment 3: Downstream Modeling}
\label{sec:Exp4}

\begin{figure*}[tbh]
    \centering
    \includegraphics[width=0.98\linewidth]{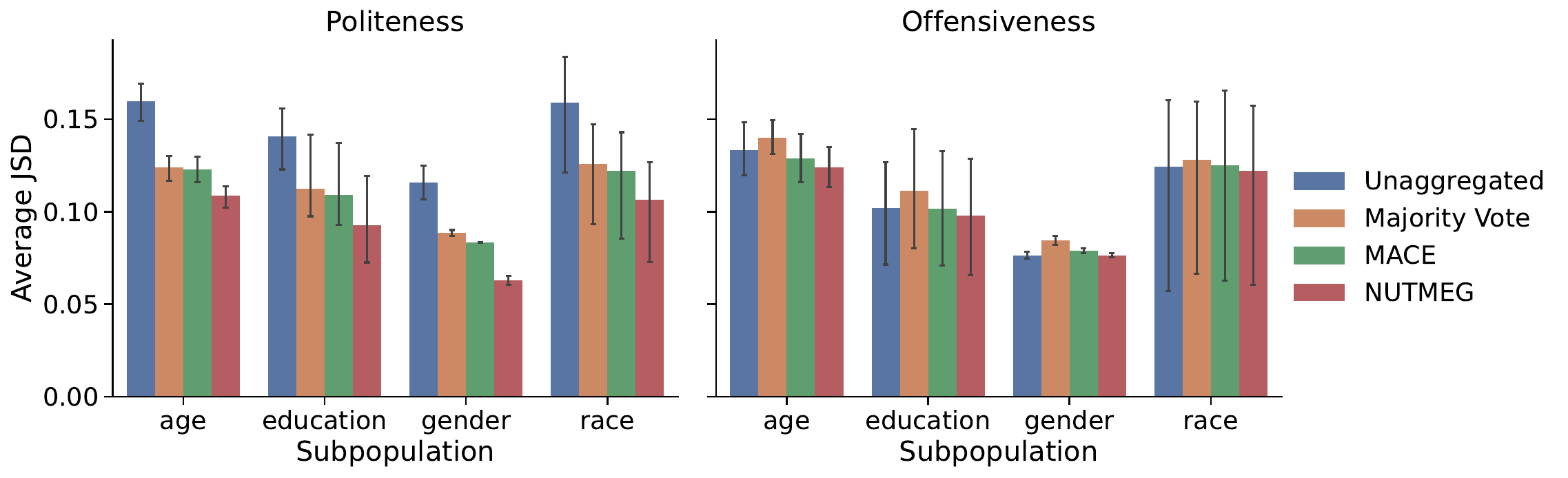}
    \caption{Performance at replicating the ground truth label distribution by subgroup (lower is better) using learning from disagreement models trained with different types of label aggregation. The bar height indicates the mean JSD across all subpopulations' label distribution in the test set, and error bars indicate the total range of JSD across all subpopulations. Our results show that any type of aggregation is helpful to reduce noise, but by jointly modeling subpopulation-preference with annotator competence, \nutmeg is better able to match the true label distributions. }
    \label{fig:lfd-results}
\end{figure*}

The purpose of \nutmeg is to remove spam annotations in human-labeled datasets while retaining valid subpopulation-level disagreement. Experiments 1 and 2 have demonstrated this on synthetic data and, here, we evaluate whether \nutmeg-aggregated labels  improve downstream modeling.

\nutmeg produces estimated true labels for each subpopulation and therefore, for predictive modeling, we adopt the learning from disagreement setting where a classifier is trained on multiple labels per annotation. We hypothesize that using the full distribution of labels (directly from annotators) introduces unnecessary noise, and in this experiment aim to answer the question: {does removing spam annotations improve performance on downstream tasks using real-world data}?

\subsection{Experimental Setup}

\myparagraph{Dataset}
To effectively use \nutmeg, we require annotated data with metadata indicating which annotators belong to which subgroups. While \nutmeg can be used for any definition of a subgroup, here, we opt to use demographics as a way to partition annotators, as multiple works have noted meaningful variation by race and gender \citep{larimore-etal-2021-reconsidering, sap-etal-2022-annotators, pei-jurgens-2023-annotator, wan_everyones_2023}. 
We use data from \textsc{Popquorn}  \citep{pei-jurgens-2023-annotator} which provides  annotations on a 1–5 Likert scale for two classification tasks (i) offensiveness and (ii) politeness, and age, race, and education demographics of annotators. We binarize the Likert ratings in their data at $\ge$3. We also remove subpopulations with $\le$5\% of annotations to ensure that the models are being trained with sufficient representation.

We note that though we choose to evaluate on data using annotator demographics to split subpopulations, \nutmeg does not necessarily require additional data collection as recent work has demonstrated the efficacy of clustering annotators into subpopulations based on annotator behaviors \citep{vitsakis-etal-2024-voices}. 

\myparagraph{Modeling}
For each task, we train a multi-task classification model, where a base model has separate classification heads for each subpopulation. This setup uses multiple tasks to represent salient disagreements and is a popular approach in learning from disagreement \citep[e.g.,][]{fornaciari-etal-2021-beyond, davani-etal-2022-dealing, mokhberian-etal-2024-capturing,  wang-plank-2023-actor}. Prior to training, we run \nutmeg on the training and validation sets using separate runs for each demographic category (i.e., gender, age, race, and education). We then fine-tune ModernBERT \citep{warner_modernbert_2024} models with additional classification layers for each subpopulation in a demographic (e.g., Man and Woman in gender). In this case, our multiple tasks are predicting the true subpopulation labels. 

As a baseline for standard aggregation methods, we also fine-tune single-task ModernBERT models on labels aggregated by either majority vote or \mace.\footnote{Because it is commonly used in NLP and performed similarly to LFC and BCC in Experiment 1, we use only \mace as a baseline for simplicity.} Finally as a baseline for training on the full distribution of disaggregated annotations, we follow the popular approach of training a multi-task model where each task is a different annotator in the dataset \citep{davani-etal-2022-dealing, wang-plank-2023-actor, mokhberian-etal-2024-capturing}. The final prediction for each subpopulation is then the average of the predicted probabilities for annotators in that subpopulation. Additional details on the training procedure are  in Appendix \ref{sec:modeling_details}. 

\myparagraph{Evaluation}
Recognizing the many causes of human label variation \citep{plank-2022-problem}, we measure model performance by comparing the predicted probabilities output by the model for each subpopulation (if it is multi-task) to the true distribution of labels provided by that subpopulation in the test set; in the single task setups, we use the same probability for each group. Following previous work, we quantify the similarity of these distributions with Jensen–Shannon divergence \citep[JSD; ][]{uma_learning_2021}, where lower is better. Note that the set of annotators in the training and validation sets is entirely separate from the set of annotators in the test set, and we score using the label distribution of the full, disaggregated labels in the test set.

\subsection{Results}

\myparagraph{Politeness} We find that for politeness detection, models trained on \nutmeg outperform both models trained on traditionally aggregated annotations and models trained on disaggregated annotations (\fref{fig:lfd-results}, left).
By learning from more-accurate aggregations that both reduce noise and highlight systematic disagreement, models are better able to predict the full label distribution for each item in the test set.
Our results show  aggregation generally helps, with even the naive but commonly-used majority voting often reducing noise in the data. Yet, the gap between \mace and \nutmeg highlights that there is additional benefit to modeling subpopulation-variation---and that for this politeness task, there is likely meaningful variation that can be modeled.
This trend supports our hypothesis that learning directly from disaggregated annotations can introduce noise and \nutmeg's noise reduction improves performance on downstream tasks. It also demonstrates that accounting for subpopulation-level variation when predicting annotator competence improves performance on all subpopulations.

\myparagraph{Offensiveness} We find that for offensiveness detection (\fref{fig:lfd-results}, right), there are no significant differences between models trained with different aggregation methods or no aggregation. However, among the aggregation steps \nutmeg does as well as or better than no-aggregation. This neutral behavior highlights it is at least not introducing additional noise, unlike majority vote which generally has higher JSD. 

We interpret this trend as an example where there is no systematic variation by subpopulation in the data. After a manual review of the data, we found that the majority of examples exhibit high subjectivity. For example, some annotators may label an encouraging statement as offensive if it contains vulgar language, while others may label an unkind statement as inoffensive if there is implied sarcasm.

\section{Conclusion}

In this work, we introduce \nutmeg, a Bayesian model that infers ground truth labels from annotations while accounting for systematic differences among annotator subpopulations. By extending item-response models, \nutmeg jointly estimates annotator competence and identifies when groups consistently diverge in their labeling decisions, addressing the limitations of traditional aggregation models that treat deviations as errors. Our experiments on synthetic and real-world data demonstrate that \nutmeg effectively recovers distinct subpopulation labels, mitigates spam annotations, and improves the performance of Learning from Disagreement models. By preserving meaningful disagreement, \nutmeg provides a more nuanced understanding of annotation data, particularly in subjective NLP tasks. Its data efficiency makes it well-suited for crowdsourcing settings, and its ability to model annotator variation contributes to more representative NLP models. We release \nutmeg and the accompanying synthetic data generation and evaluation libraries. This work highlights the importance of incorporating diverse perspectives in annotation modeling and encourages further research into principled approaches for handling disagreement in human-labeled data.

\section {Limitations}

\nutmeg works by identifying systematic disagreement in subpopulations of annotators. While we have demonstrated that \nutmeg can correctly identify such disagreement, most real data contains label variation beyond that due to subpopulations. Thus, while the method can potentially improve quality, it is not a universal panacea for noisy annotations. Indeed, we note that  \nutmeg doesn't always improve downstream model performance (in the Offensiveness task) suggesting that even when demographic labels are present and modeled, other sources of label variation in the data may more strongly influence performance; indeed, past work in NLP has found a mixed trend, where for some datasets, demographics explain little variation \citep{Orlikowski2023TheEF}, while for others, demographics explain substantial variation \citep{larimore-etal-2021-reconsidering, sap-etal-2022-annotators, pei-jurgens-2023-annotator, wan_everyones_2023}.

\nutmeg produces estimates of the ground truth by subpopulation. We anticipate that these labels will be most useful for practitioners who use learning from disagreement models that are designed to model subpopulations. However, most of NLP still uses models that produce a single label. Our work could still potentially help in these settings by allowing practitioners to train models for separate populations (e.g., an offensive language detector optimized for the views of a particular subpopulation) and then deploy these strategically. Further, even when a traditional machine learning model is used, \nutmeg can still help identify meaningful disagreement in the data and raise awareness for the practitioner.

Our experiments with synthetic and real data used a known, discrete subpopulation label for each annotator. However, for many datasets, this type annotator information is not present. While we note that recent work has pointed to the ability to cluster annotators to create inferred subpopulations \citep{vitsakis-etal-2024-voices}, we have not evaluated that strategy here in favor of first demonstrating that the method works with known demographics. Our experimental design limits the potential confounding influence of the clustering model on the potential benefits of \nutmeg. Nevertheless, we view this as a promising direction for future work to explore.

Finally, our experiments used a limited number of real-world datasets to demonstrate effectiveness. Demographically-labeled data is growing in NLP \citep[cf.][]{Santy2023NLPositionalityCD}, but still uncommon. The \textsc{Popquorn} dataset is among the largest available dataset with multiple tasks, making it ideal for our study. However, we recognize that future work could evaluate \nutmeg with more datasets as they become available. 

\section{Ethical Considerations}

\nutmeg requires that annotators be associated with a particular subpopulation. We anticipate that for many practitioners, this background will be based on demographics or other personal attributes. As a result, \nutmeg could potentially increase the collection of personal data, which needs to be responsibly stored. However, we view this risk as being outweighed by the benefits of having the different views of subpopulations better represented in models.

\bibliography{anthology,custom}

\clearpage

\appendix

\section{Modeling Details} \label{sec:modeling_details}
\nutmeg runs entirely on CPU and can be run on any reasonably equipped computer.

We trained the ModernBERT models with 149M parameters on a single NVIDIA RTX A6000 GPU  Hugging Face Transformers 4.48.1 \cite{wolf-etal-2020-transformers} and PyTorch 2.5.1 \cite{NEURIPS2019_9015} on a CUDA 12.4 environment. All models were trained for 12 epochs with a batch size of 192 and were tuned for a learning rate in the range $[1 \times 10^{-5}, 2 \times 10^{-3}]$ with Optuna \citep{akiba_optuna_2019}. We use the train, validation, and test splits provided by the original \textsc{Popquorn} dataset \citep{pei-jurgens-2023-annotator}. To ensure reproducibility, we set all random seeds in Python to be 42. Full model parameters are available at {\small{http://anon}}.

\end{document}